\documentclass[lettersize,journal]{IEEEtran}
\usepackage{graphicx}
\usepackage{lipsum}
\usepackage{soul}
\usepackage{xcolor}
\sethlcolor{yellow}
\usepackage[caption=false,font=normalsize,labelfont=sf,textfont=sf]{subfig}
\usepackage{booktabs}

\begin{document}
\title{Environmental Feature Engineering and Statistical Validation for ML-Based Path Loss Prediction}
\author{Jonathan~Ethier, Mathieu~Châteauvert, Ryan~G.~Dempsey, and Alexis~Bose
\thanks{Submitted to IEEE Antennas and Propagation Wireless Letters (AWPL), August 2025. All authors are with the Communications Research Centre, Kanata, Ontario, Canada,  e-mail: jonathan.ethier, mathieu.chateauvert, ryan.dempsey, alexis.bose @ ised-isde.gc.ca}}
\markboth{Journal of IEEE Antennas and Wireless Propagation Letters,~Vol.~XX, No.~XX, Month~Year}
{Author1 \MakeLowercase{\textit{et al.}}: Your Paper Title}
\maketitle

\begin{abstract}
Wireless communications rely on path loss modeling, which is most effective when it includes the physical details of the propagation environment. Acquiring this data has historically been challenging, but geographic information systems data is becoming increasingly available with higher resolution and accuracy. Access to such details enables propagation models to more accurately predict coverage and account for interference in wireless deployments. Machine learning-based modeling can significantly support this effort, with feature-based approaches allowing for accurate, efficient, and scalable propagation modeling. Building on previous work, we introduce an extended set of features that improves prediction accuracy while, most importantly, proving model generalization through rigorous statistical assessment and the use of test set holdouts.
\end{abstract}


\section{Introduction}
Propagation modeling is essential for enhancing communication networks. With the wide range of wireless technologies, from IoT devices \cite{IoT}, mature Fifth-Generation (5G) networks \cite{Later5G}, and the rise of Sixth-Generation (6G) \cite{6G}, accurate propagation models will remain relevant as wireless networks evolve. These models predict radio wave behavior in a variety of environments, ensuring appropriate coverage, acceptable interference levels, and efficient use of spectrum. As seamless communication becomes the standard, fast and accurate propagation models will continue to grow in importance.

Our prior work \cite{ethierAWPL1} considered a lean set of features (frequency, distance, and total obstruction depth) to make accurate path loss predictions. This letter makes three new contributions: (1) we more than double the number of features used in the ML model, improving model accuracy while further improving the generalized behavior of the model, (2) we conduct rigorous statistical studies of the model sensitivity to initialization and train/validation splits, and (3) we introduce additional blind test sets to further confirm that the richer feature set does not lead to overfitting, which can be a concern in feature-rich models \cite{bishop2006pattern}. In addition to these three new contributions, our ML model is distinct from other modeling approaches in that it strictly uses generic obstructions for feature generation that are easily acquired worldwide. This work does not require access to more complex datasets that differentiate terrain from clutter, nor does it require land use details. Future work could explore if higher complexity environment details \cite{reviewer1_1of2,reviewer1_2of2} could result in better performance.

This work is related to, though distinct from \cite{ryanAWPL}, where the goal in that letter was to construct path-specific path loss models, using the entire path profile as an input to the path loss model. The authors in \cite{MinFeatures} consider a set of features that do not directly connect to physical Geographic Information Systems (GIS) features, as we have done here. In \cite{pathloss_manyfeatures}, the authors develop feature-rich, region-specific models and use cross-validation with no holdouts to assess model performance. In both \cite{ethierAWPL1} and \cite{bocusIEEEaccess}, the authors use test holdouts from the same country used in the training data, and neither performs statistical studies of the model performance. To assess model generalization rigorously \cite{gis-split}, we use both statistical studies and intercontinental blind test holdouts.


\section{Proposed Method}
\label{sec:propmethod}

\subsection{Data and Preprocessing}
\label{subsec:datasets}
The training data used in this work is radio frequency (RF) drive test data \cite{OFCOM_DATA_OPEN} collected by the UK's Office of Communications (Ofcom). Links are highly obstructed with distances that range from 0.25 to 50 km, antenna heights from 17 to 25 m, a fixed receiver height of 2 m, in urban, suburban and rural environments. More details about these measurements can be found in \cite{ethierAWPL1} and \cite{bocusIEEEaccess}. Path profiles can be extracted from UK Open Data \cite{UK_DTM_DSM}, using supporting code from \cite{SAFE_GITHUB}. The Digital Terrain Model (DTM) and Digital Surface Model (DSM) are extracted from the GIS data. All subsequent features are derived from these sources of information, only using samples above the measurement noise floor with an additional 6~dB margin. A total of 20\,000 random samples meeting this noise criterion were extracted for each of the six measurement frequencies (449, 915, 1802, 2695, 3602, and 5850 MHz) within each of the six drive tests (London, Merthyr Tydfil, Nottingham, Southampton, Stevenage, Boston), resulting in a training set size of 720\,000 samples (20\,000 samples x 6 frequencies x 6 drive test locations).

Additional proprietary drive test data was acquired from Netscout Systems Inc. \cite{NetScout}. The dataset consists of over 120\,000 measurements from ten regions across Canada, with measurements around 750, 1950, 2600 and 3650 MHz, using similar ranges of antenna heights,  link distances, and environment types as the UK data. Path profiles and their features were all acquired in the same manner as the UK drive tests, using the High-Resolution Digital Elevation Model (HRDEM) \cite{HRDEM} for the GIS information (DTM and DSM) because of the Canadian geographic locations. Earth curvature is accounted for when extracting the DTM and DSM path profiles, using the mean radius of 6 371 km.

\subsection{Model Features: Depth, Density, and Distance}
\label{subsec:features}
\begin{figure*}[!t]
\centering
\includegraphics[width=0.99\textwidth]{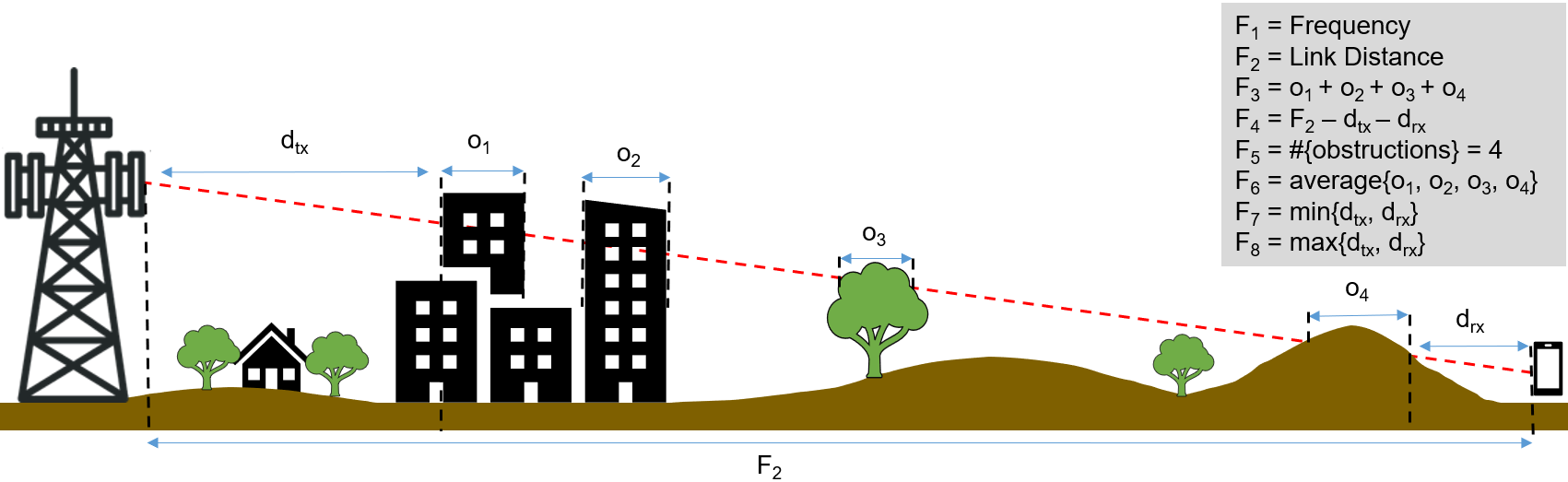}
\caption{Path profile with a mixture of buildings (black), terrain (brown) and foliage (green) obstructions with the direct path between transmitter and receiver shown in red. Since the model uses DSM-only to assess obstructions, all obstructions are treated identically, and only DSM statistics matter when computing the eight model features. All depths and distances are measured along the direct path and take into account both angle relative to ground and Earth curvature.}
\label{fig:feature_extraction}
\end{figure*}

The features proposed are all scalar quantities and are derived from the intersection of the direct path (the line connecting the transmitter to the receiver) and DSM. Fig. \ref{fig:feature_extraction} shows the extraction of all eight features from an example path profile, and Table \ref{tab:feature_descriptions} lists the eight features. Beyond model performance, the choice of features is justified by their assessment of obstructions along the direct path through foundational physics metrics such as \textit{distance}, \textit{depth} and \textit{density}.

\subsubsection{Fundamental Features}
The two fundamental features present in most propagation models are \( F_1\) \textit{frequency} and \( F_2\) \textit{distance from the transmitter to the receiver}. This enables the ML model to learn the path loss exponent and improve upon the basic free-space path loss.

\subsubsection{Obstruction Depth Features}
Consider the two depth-based features: \( F_3 \) \textit{total obstruction depth} and \( F_4 \) \textit{distance from first to last obstruction}. These \textit{depth} features provide the amount of the direct path that is obstructed from transmitter to receiver, as well as the extent over which obstructions exist, which can be predictive of the attenuation along the link. Feature four is similar to the metric introduced in \cite{KozmaEuCAP}.

\subsubsection{Obstruction Density Features}
A block is defined as a contiguous obstruction along the direct path, with no blockage before and after the block. The total number of blocks along the link (an integer count) is represented with \( F_5 \), and feature \(F_6\) is the average depth of these blocks. This provides the model with \textit{density} information of the obstructions, which can potentially help to implicitly model multi-path along the link.

\subsubsection{Obstruction Distance Features}
Lastly, we use the minimum and maximum distances from Tx and Rx to obstructions, \( F_7 \) and \( F_8 \), respectively. These features provide the ML model the relative location of obstructions along the link.

An exhaustive search of metrics such as min, max, median, skew and SD was considered for all features, with the resulting eight metrics shown to be optimal for improving validation scores. Additional features associated with depth, density and distance were considered, but showed diminishing returns. Future work can consider additional height-based metrics associated with diffraction. All eight features are identical when viewed from the Tx to Rx or Rx to Tx, i.e. they are reciprocal features. This is important when constructing a general path loss model \cite{RyanTurkey}.

\begin{table}[ht]
\centering
\caption{Feature Descriptions}
\begin{tabular}{lll}
\toprule
\textbf{Symbol} & \textbf{Feature} & \textbf{Units} \\
\midrule
\( F_1 \) & Frequency & MHz \\
\( F_2 \) & Distance from Transmitter to Receiver & Meters \\
\( F_3 \) & Total Obstruction Depth & Meters \\
\( F_4 \) & Distance from First to Last Obstruction & Meters \\
\( F_5 \) & Number of Contiguous Blocks & Count\\
\( F_6 \) & Average Block Depth & Meters \\
\( F_7 \) & Minimum Distance from Tx/Rx to Obstructions & Meters\\
\( F_8 \) & Maximum Distance from Tx/Rx to Obstructions  & Meters \\
\bottomrule
\end{tabular}
\label{tab:feature_descriptions}
\end{table}
             
\subsection{Model Architecture}
Given tabular features, dense neural networks are an appropriate model architecture \cite{Goodfellow-et-al-2016} for modeling. The number of hidden layers and the number of neurons per layer were optimized based on validation scores with the ideal structure of two hidden layers and 64 neurons per layer, resulting in at most 4,801 total parameters. Additional layers (3, 4, 5) and/or neurons (128, 256, 512) led to no improvement or overfitting and using a single layer and/or fewer than 64 neurons resulted in underfitting. Both extremes resulted in the validation MSE remaining the same or increasing. A dropout layer was included after every hidden layer during training. Rectified linear unit (ReLU) activations were used to exploit non-linear interactions between features, except for the output layer, which used a linear activation.

We will perform an ``add two'' ablation study of three configurations of model inputs that follow increasing model complexity, as summarized in Table \ref{tab:feature_summary}.

\begin{table}[ht]
\centering
\caption{Feature Configurations for ``add two'' ablation study using the Eight Features from Section \ref{subsec:features}}
\begin{tabular}{ll}
\toprule
\textbf{Configuration} & \textbf{Features} \\
\midrule
4 Features & \( F_1 \), \( F_2 \), \( F_3 \), \( F_4 \) \\
6 Features & 4 Features + \( F_5 \), \( F_6 \) \\
8 Features & 6 Features + \( F_7 \), \( F_8 \) \\
\bottomrule
\end{tabular}
\label{tab:feature_summary}
\end{table}

\subsection{Training Approach, UK and Canadian Tests}

Six holdout scenarios are constructed, each with one drive test held out of training, and the remaining five drive tests forming the train and validation data. This provides six proper geographically (and morphologically) distinct test sets. Every test scenario is run 20 times independently, with random starting model weight and training splits. The results of the 20 runs provide a mean and standard deviation (SD) of the root mean squared error (RMSE), allowing the performance and variation of the models to be judged.
 
Hyperparameters include: (a) random train and validation split of 80\% / 20\%, (b) batch size of 8192, (c) dropout of 25\%, (d) Adaptive moment estimation (Adam) optimizer with an initial learning rate of 0.001, and (e) early-stopping patience of 50 epochs. Mean squared error (MSE) is used as the loss function to penalize outliers more than mean absolute error (MAE) due to the squared error term in the former. Minimzing MSE also has the additional benefit of directly reducing the variance of error \cite{bishop2006pattern}. MSE is preferred in RF propagation work since accurately modeling outliers is important in interference studies. Additionally, MSE (and RMSE) are used over normalized MSE (NMSE) since retaining the units of prediction is important to compare against other path loss models.

The use of batch and layer norms was investigated, but they were shown not to improve validation scores and were therefore not used in the final architecture. Each input feature is normalized to have zero mean ($\mu$) and SD ($\sigma$) of 1, and only training samples are used to determine the normalization. 

For eight features, optimization runs require, on average, 90 epochs and 2.3 minutes to converge, using an m6i.2xlarge Amazon Web Service (AWS) instance.

An additional train, validation and test scenario is considered using identical hyperparameters and holdouts and an additional ``no holdout'' scenario (training and validating on all UK drive tests) and testing on Canadian drive test data (see \ref{subsec:datasets}). The results of this approach are found in Section \ref{subsec:modelperformance_can}.


\section{Results and Discussion}
\label{sec:results}

\subsection{Model Performance, UK Blind Tests}
\label{subsec:modelperformance_uk}

A summary of the mean and SD RMSE test scores for the six drive test holdouts (discussed in Section \ref{subsec:datasets}) is presented in Table \ref{tab:feature_metrics_uk}, along with a comparison to P.1812 \cite{P1812}, which uses the same GIS data source as the ML model for path loss predictions. In every holdout, the test RMSE decreases with the inclusion of additional features, as further supported by the low SD of the scores. The mean RMSE across the six holdouts decreases steadily to as low as 7.0 dB for eight features and, in all cases, remains lower than that of P.1812. Adding the density features \(F_5\) and \(F_6\) and the distance-to-obstruction features \(F_7\) and \(F_8\) improves test scores in every holdout and, on average, yields statistically significant reductions in RMSE of 0.3 and 0.1 dB, respectively. Interestingly, two of the holdouts (Merthyr Tydfil and Stevenage) exhibit notably higher mean RMSEs than the other four holdouts. This is likely due to Merthyr Tydfil consisting of hilly, terrain-dominant links and Stevenage consisting of more modern building infrastructure, making both holdouts challenging for model generalization due to their distinct characteristics relative to the other drive tests.

\begin{table}[ht]
\centering
\caption{Mean and SD of Model RMSE on UK Holdout Drive Test Data}
\begin{tabular}{lccccccc}
\toprule
 & \textbf{P.1812} & \multicolumn{2}{c}{\textbf{4 Features}} & \multicolumn{2}{c}{\textbf{6 Features}} & \multicolumn{2}{c}{\textbf{8 Features}} \\
\textbf{Holdout} & ref. \cite{ethierAWPL1} & Mean & SD & Mean & SD & Mean & SD \\
\midrule
London         & 8.8  & 6.9 & 0.1 & 6.8 & 0.1 & \textbf{6.7} & 0.1 \\
Merthyr Tydfil & 13.4 & 7.9 & 0.1 & 7.6 & 0.1 & \textbf{7.6} & 0.2 \\
Nottingham     & 12.6 & 6.9 & 0.1 & 6.9 & 0.2 & \textbf{6.7} & 0.1 \\
Southampton    & 9.5  & 6.7 & 0.1 & 6.4 & 0.1 & \textbf{6.3} & 0.1 \\
Stevenage      & 12.3 & 8.5 & 0.1 & 7.9 & 0.1 & \textbf{7.8} & 0.1 \\
Boston         & 11.4 & 7.2 & 0.3 & 7.1 & 0.1 & \textbf{6.7} & 0.2 \\
\midrule
Mean       & 11.3 & 7.4 & 0.1 & 7.1 & 0.1 & \textbf{7.0} & 0.1 \\
\bottomrule
\end{tabular}
\label{tab:feature_metrics_uk}
\end{table}

\subsection{Model Performance, Canadian Blind Tests}
\label{subsec:modelperformance_can}

The results of the blind test holdouts shown in Section \ref{subsec:modelperformance_uk} provide strong evidence for model generalization. However, since the measurements were performed in the same country, by the same measurement team, there can remain some lingering doubts regarding generalization. To address this potential concern, we investigate model performance on an additional set of Canadian drive tests discussed in Section \ref{subsec:datasets}.

\begin{figure}[ht]
\begin{center}
\noindent
  \includegraphics[width=3.0in]{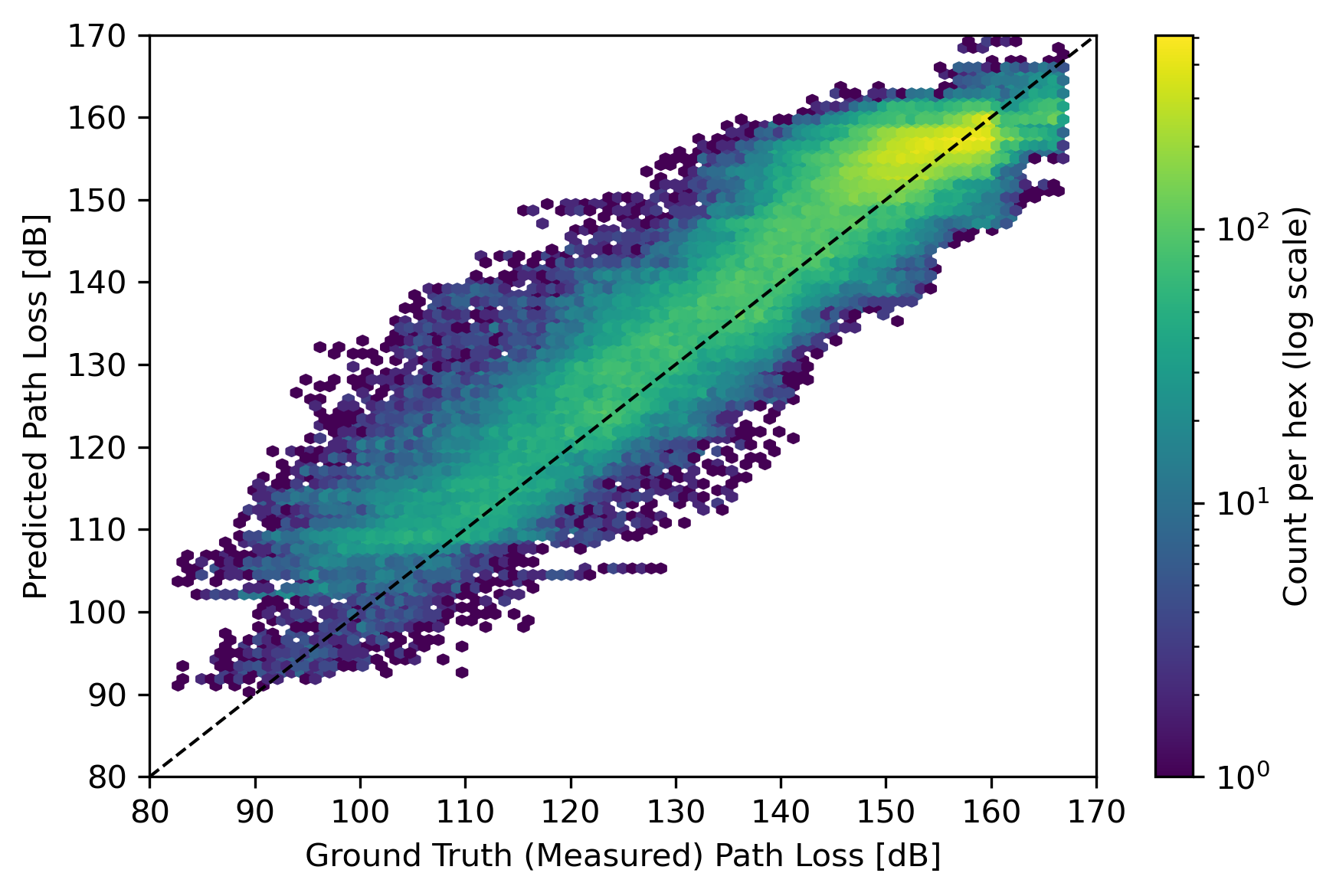}
  \caption{Hexagonal-binning (2-D histogram) plot of predicted vs. measured path loss; colour indicates samples per bin (log scale). Model is trained on all UK drive tests, blind test on Canadian drive tests and has a coefficient of determination $R^{2}$ equal to 0.88 and an RMSE of 6.74 dB.}
  \label{fig:scatterplotresults_can}
\end{center}
\end{figure}

In our seventh model labelled ``no holdout'', we use all six UK drive tests for training and validation, and test on the Canadian drive tests. This provides model training with the complete set of UK drive tests while using the Canadian drive tests for blind assessment. The results of the seven models are summarized in Table \ref{tab:feature_metrics_can}. A hexagonal binned 2-D histogram plot (an alternative to scatter plots that addresses overplotting in large datasets) of the predicted vs. measured test data for the eight-feature model is shown in Fig. \ref{fig:scatterplotresults_can}.

Note that the eight-feature model has the lowest RMSE for all six holdouts and the seventh ``no holdout'' model. Some observations include: (1) the Boston holdout has the highest test RMSE, indicating the importance of having the Boston drive test data in training, and (2) the Merthyr Tydfil holdout model does not suffer degraded performance when tested on the Canadian data since Merthyr Tydfil consists of hilly, terrain-dominant drive tests, which were not present in the Canadian test data.

Since the eight-feature model shows consistently low mean RMSE for all train, validation and test configurations, we conclude that this is a well-generalized model. Given the low SD of the mean RMSE ($\leq$0.2 dB), model selection is not highly sensitive, as most models are likely to exhibit consistently good performance. The resulting model performance shows improvements over previous work \cite{ethierAWPL1} in a statistically significant manner (as per the SD of RMSE). Additionally, the 6.3 dB blind test RMSE on Southampton is lower than the 7.3 dB RMSE on the same test set in \cite{bocusIEEEaccess}. The ``no holdout'' model blind test RMSE (last row) in Table \ref{tab:feature_metrics_can} shows a clear ``add two'' stepwise trend towards lower RMSE, statistically significant generalized modeling performance.

\begin{table}[ht]
\centering
\caption{Mean and SD of Model RMSE on Blind Canadian Test Data}
\begin{tabular}{lcccccc}
\toprule
 & \multicolumn{2}{c}{\textbf{4 Features}} & \multicolumn{2}{c}{\textbf{6 Features}} & \multicolumn{2}{c}{\textbf{8 Features}} \\
\textbf{Holdout}
 & Mean & SD & Mean & SD & Mean & SD \\
\midrule
London         & 7.4 & 0.2 & 7.2 & 0.1 & \textbf{7.0} & 0.1 \\
Merthyr Tydfil & 7.1 & 0.1 & 6.8 & 0.1 & \textbf{6.8} & 0.1 \\
Nottingham     & 7.1 & 0.1 & 7.0 & 0.2 & \textbf{6.9} & 0.2 \\
Southampton    & 7.0 & 0.1 & 6.8 & 0.1 & \textbf{6.7} & 0.1 \\
Stevenage      & 6.9 & 0.1 & 6.8 & 0.1 & \textbf{6.8} & 0.2 \\
Boston         & 7.4 & 0.1 & 7.2 & 0.2 & \textbf{7.1} & 0.1 \\
\midrule
Mean       & 7.1 & 0.1 & 7.0 & 0.1 & \textbf{6.9} & 0.2 \\
\midrule \midrule
No Holdout   & 7.1 & 0.1 & 6.9 & 0.1 & \textbf{6.7} & 0.2 \\
\bottomrule
\end{tabular}
\label{tab:feature_metrics_can}
\end{table}

\subsection{Assessing Risk of Overfitting}
\label{subsec:deepdive}
In Section \ref{subsec:modelperformance_uk} and Section \ref{subsec:modelperformance_can}, we provided evidence for robust models since they all exhibit low variation of RMSE over 20 train/validation splits for both inter-country and inter-continental holdouts. We further assess this claim and run the optimization of the eight-feature model 200 times with random initialization and train/validation splits, resulting in the statistics summarized in Table \ref{tab:longer_run}. If we specifically consider the 20 best models based on the validation scores from the 200 optimization runs, a mean test RMSE of \( 6.75 \pm 0.08 \) dB is achieved. Additionally, the poorest test score of 7.63 dB comes from the model with the poorest validation score of 6.97 dB. These results show that one can consistently achieve superior test scores by selecting models based on the validation score, as this approach does not lead to overfitting.

\begin{table}[ht]
\centering
\caption{RMSE Statistics for Longer 200 Optimization Runs}\begin{tabular}{lcc}
\toprule
\textbf{RMSE} & \textbf{UK (Val)} & \textbf{Canadian (Test)} \\
\midrule
Min     & 6.31 & 6.49 \\
Max     & 6.97 & 7.63 \\
Median  & 6.47 & 6.77 \\
Mean (200 models)    & \(6.50 \pm 0.12 \) & \(6.79 \pm 0.18\) \\
Mean (Best 20 models)   & \(6.37 \pm 0.03 \) & \(6.75 \pm 0.08\) \\
\bottomrule
\end{tabular}
\label{tab:longer_run}
\end{table}


\subsection{Frequency- and Distance-Dependent Performance}

\begin{figure}[ht]
\begin{center}
\noindent
  \includegraphics[width=3.2in]{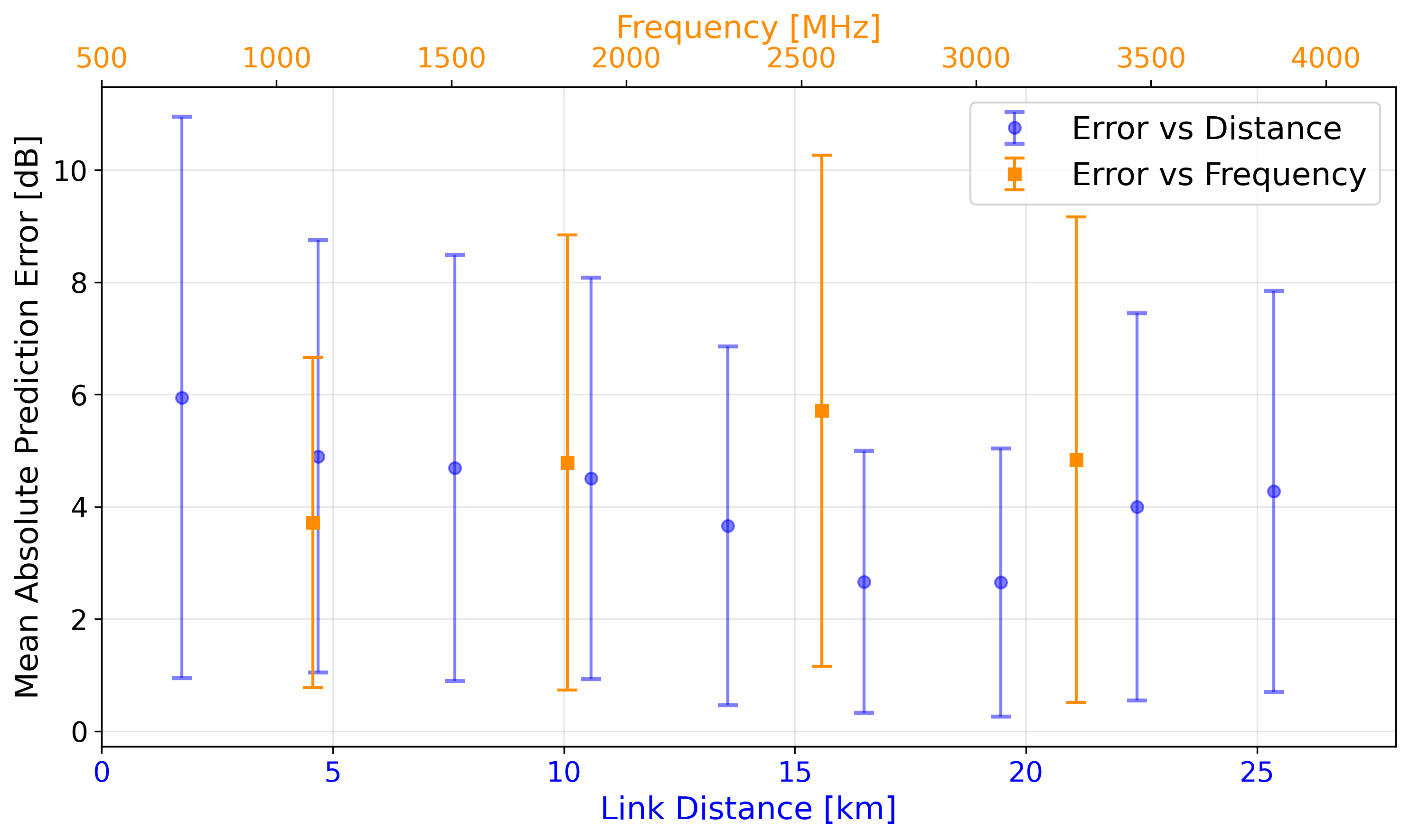}
  \caption{Mean absolute error on Canadian test data (whiskers for SD of error) as a function of link distance (3 km bins, $>$10000 samples per bin, bottom x-axis) and frequency (no binning, four discrete frequencies, top x-axis)}
  \label{fig:freq_and_dist}
\end{center}
\end{figure}

In Section \ref{subsec:datasets}, we discussed the link distances and frequency ranges of the data sets. While the frequency ranges overlap, the specific frequencies used in the Canadian test data differ from the UK data. This allows us to evaluate the model's interpolation ability within the 0.5--6 GHz domain, the range over which we expect our model to generalize. In Fig. \ref{fig:freq_and_dist}, we show the mean absolute prediction error on the Canadian blind test data as a function of link distance and frequency. 

We observe a slight decrease in prediction error for longer link distances $(r\approx-0.1$, where $r$ is the Pearson coefficient), consistent with the higher proportion of longer links in the training data \cite{OFCOM_DATA_OPEN}, which enables more accurate path loss prediction at longer ranges. Improving performance for shorter links may require increasing the number of training samples in that distance range.

Additionally, prediction error increases slightly with frequency $(r\approx+0.1)$, which is sensible given that electromagnetic waves interact with smaller objects at higher frequencies and become increasingly sensitive to the material properties of obstructions. To extend the model towards frequencies greater than 6 GHz, new features such as identifying the material properties of obstructions may need to be introduced to maintain the low prediction errors and model generalization.

Overall, despite these small correlations, the prediction errors are low across link distances and frequencies, confirming that the model generalizes well across the 0.5--6 GHz band.

\section{Concluding Remarks}
\label{sec:conc}

This letter described the use of scalar features to model path loss from 500 MHz to 6 GHz. An optimum set of eight scalar features was used, offering low generalized prediction error, achieving less than 7 dB RMSE on intercontinental blind testing. Additional studies showed that the test performance does not suffer from overfitting when relying on the validation scores, and that the modeling is consistent across a wide range of link distances and frequencies. The model architecture is simple yet effective, and the feature extraction process from path profiles is straightforward, requiring only surface (DSM) information. Future work can involve incorporating additional features to better account for diffraction effects.

\bstctlcite{IEEEexample:BSTcontrol}

\end{document}